\documentclass{llncs}

\usepackage{silence}
\usepackage[english]{babel}
\usepackage[utf8]{inputenc}
\usepackage[T1]{fontenc}
\usepackage{amsmath,amssymb}
\usepackage{mathtools}
\usepackage{fca}
\usepackage[english,style=iso]{datetime2}
\usepackage{listings}
\usepackage[kerning=true, protrusion=true,babel]{microtype}
\usepackage[giveninits=true,mincrossrefs=5]{biblatex}
\usepackage[hyperfootnotes=false]{hyperref}
\usepackage[capitalise]{cleveref}
\renewcommand{\epsilon}{\varepsilon}
\renewcommand{\phi}{\varphi}

\newcommand{\ie}{i.e.,}
\newcommand{\eg}{e.g.,}

\usepackage{csquotes}
\usepackage{booktabs}
\usepackage{paralist}
\usepackage{todonotes}
\presetkeys{todonotes}{color=blue!5}{}

\mathtoolsset{centercolon}
\Crefname{problem}{Problem}{Problems}
\providecommand\given{}
\makeatletter
\newcommand\Set@Symbol[1][]{%
\nonscript\:#1\vert{}%
\allowbreak{}%
\nonscript\:%
\mathopen{}}%
\DeclarePairedDelimiterX{\set}[1]{\lbrace}{\rbrace}{%
\renewcommand{\given}{\Set@Symbol[\delimsize]}%
#1%
}
\makeatother

\DeclarePairedDelimiterXPP{\W}[1]{\mathcal{W}}{\lparen{}}{\rparen}{}{#1}
\newcommand{\WD}[2]{#1 (\enquote{#2})} % wikidata entities
\newcommand{\WDM}[2]{\ensuremath{{\texttt{#2}_{#1}}}}
\DeclarePairedDelimiter{\abs}{\lvert{}}{\rvert{}}
\DeclarePairedDelimiter{\tuple}{\langle{}}{\rangle{}}
\DeclarePairedDelimiterXPP{\Family}[2]{}{\lparen{}}{\rparen{}}{_{#2}}{#1}
\DeclarePairedDelimiterXPP{\powerset}[1]{\mathfrak{P}}{\lparen{}}{\rparen{}}{}{#1}
\DeclarePairedDelimiterXPP{\finitepowerset}[1]{\mathfrak{P}}{\lparen{}}{\rparen{}}{_\text{fin}}{#1}

\DeclarePairedDelimiter{\statementBody}{\lparen{}}{\rparen{}}
\newcommand{\qualifierValue}{\mathpunct{:}}

\DeclareMathOperator{\subject}{subj}
\DeclareMathOperator{\object}{obj}
\DeclareMathOperator{\annotation}{ann}
\DeclareMathOperator{\supp}{supp}
\DeclareMathOperator{\conf}{conf}
\DeclarePairedDelimiterXPP{\Qualifiers}[1]{\operatorname{Annotations}}{\lparen{}}{\rparen}{}{#1}
\DeclarePairedDelimiterXPP{\Instances}[1]{\operatorname{Classes}}{\lparen{}}{\rparen}{}{#1}
\newcommand{\plain}{\texttt{plain}}
\newcommand{\direction}{\texttt{dir}}
\newcommand{\qualifier}{\texttt{qual}}
\newcommand{\instance}{\texttt{class}}

\newcommand{\dataset}[1]{\texttt{#1}}

\newcommand{\complclass}[1]{{\sc #1}\xspace} % font for complexity classes

\newcommand{\PTime}{\complclass{P}}

\newcommand{\coNP}{\complclass{coNP}}

\newcommand{\newentity}[3]{%
\csdef{#1}{\relax\ifmmode\WDM{#2}{#3}%
\else\WD{#2}{#3}\fi}%
}
\newcommand{\inverseProperty}[1]{%
  \ifmmode{\hat{~}}%
  \else\textasciicircum%
  \fi%
  {#1}%
}

\newentity{sanFrancisco}{Q62}{San Francisco}
\newentity{planet}{Q634}{Planet}
\newentity{pluto}{Q339}{Pluto}
\newentity{frankfurt}{Q1794}{Frankfurt}
\newentity{spock}{Q16341}{Spock}
\newentity{lizTaylor}{Q34851}{Elizabeth Taylor}
\newentity{census}{Q39825}{census}
\newentity{estimation}{Q965330}{estimation}
\newentity{richardBurton}{Q151973}{Richard Burton}
\newentity{inverseF}{Q3075242}{inverse function}
\newentity{AlbertMaysles}{Q15782045}{Albert Maysles}
\newentity{itemGrandparent}{Q167918}{grandparent}
\newentity{taxonFamily}{Q35409}{family}
\newentity{taxonOrder}{Q36602}{order}
\newentity{greekDeity}{Q22989102}{Greek deity}
\newentity{human}{Q5}{human}

\newentity{propsAwards}{Q56150830}{… awards, prizes and honours}
\newentity{propsFamily}{Q22964231}{… human relationships}
\newentity{propsMath}{Q22988631}{… mathematics}
\newentity{propsSpacecraft}{Q28104992}{… spacecraft}
\newentity{propsTimeDuration}{Q51077473}{… time and duration}

\newentity{nationalMedalOfArts}{Q1789030}{National Medal of Arts}

\newentity{father}{P22}{father}
\newentity{mother}{P25}{mother}
\newentity{spouse}{P26}{spouse}
\newentity{instanceOf}{P31}{instance of}
\newentity{child}{P40}{child}
\newentity{subclassOf}{P279}{subclass of}
\newentity{determinationMethod}{P459}{determination method}
\newentity{startTime}{P580}{start time}
\newentity{endTime}{P582}{end time}
\newentity{pointInTime}{P585}{point in time}
\newentity{population}{P1082}{population}
\newentity{sibling}{P3373}{sibling}
\newentity{inputSet}{P1851}{input set}
\newentity{domain}{P1568}{domain}
\newentity{codomain}{P1571}{codomain}
\newentity{NMAW}{P5719}{National Medal of Arts winner ID}
\newentity{isNominatedFor}{\inverseProperty{P1411}}{\inverseProperty{nominated for}}
\newentity{awardReceived}{P166}{award received}

\newentity{relative}{P1038}{relative}
\newentity{typeOfKinship}{P1039}{type of kinship}
\newentity{parentTaxon}{P171}{parent taxon}
\newentity{taxonRank}{P105}{taxon rank}

\newentity{brother}{P7}{brother}
\newentity{sister}{P9}{sister}
\newentity{grandparent}{P45}{grandparent}
\newentity{order}{P70}{order}
\newentity{family}{P71}{family}
\newentity{mainTypeGND}{P107}{main type GND}
\newentity{administrativeEntity}{P132}{administrative entity}
\newentity{languageFamily}{P133}{language family}

\newentity{isMother}{\inverseProperty{P25}}{\inverseProperty{mother}}
\newentity{godparent}{P1290}{godparent}
\newentity{AngelinaJolie}{Q13909}{Angelina Jolie}
\newentity{Victoria}{Q9439}{Victoria}
\newentity{NaomiWatts}{Q132616}{Naomi Watts}
\newentity{MileyCyrus}{Q4235}{Miley Cyrus}

\newentity{nobelprizeid}{P3188}{Nobel prize ID}
\newentity{isAwardReceived}{\inverseProperty{P166}}{\inverseProperty{award received}}
\newentity{nominatedFor}{1411}{nominated for}
\newentity{catRec}{P2517}{category for recipients of this award}
\newentity{godParent}{P1290}{godparent}
\newentity{partner}{P451}{partner}
\newentity{sibling}{P3373}{sibling}
\newentity{isFather}{\inverseProperty{P22}}{\inverseProperty{father}}
\newentity{isRelative}{\inverseProperty{P1038}}{\inverseProperty{relative}}
\newentity{vFigure}{P1678}{has vertex\,\,figure}
\newentity{fPolytope}{P1678}{has facet polytope}
\newentity{base}{P3263}{base}
\newentity{isDualTo}{\inverseProperty{P1322}}{\inverseProperty{dual to}}
\newentity{mathworld}{P2812}{MathWorld\,\,identifier}
\newentity{periapsis}{P2244}{periapsis}
\newentity{typeOfOrbit}{P522}{type of orbit}
\newentity{apoapsis}{P2243}{apoapsis}
\newentity{isTypeOfOrbit}{\inverseProperty{P522}}{\inverseProperty{type of orbit}}
\newentity{orbitalPeriod}{P2146}{orbital period}
\newentity{producer}{P162}{producer}
\newentity{wdCountry}{P17}{country} % \country is used by the ACM style
\newentity{genre}{P136}{genre}
\newentity{isChild}{\inverseProperty{P40}}{\inverseProperty{child}}

\newentity{PMIDID}{P6037}{ProsopoMaths ID}
\newentity{tempRangeEnd}{P524}{temporal range end}
\newentity{tempRangeStart}{P523}{temporal range start}

\title{Discovering Implicational Knowledge in Wikidata}
\newcommand{\putorcid}[1]{\makebox[0pt][r]{\raisebox{-1ex}[0pt][0pt]{\textcolor{red}{\tiny  #1}}}}
\newcommand\blfootnote[1]{%
  \begingroup
  \renewcommand\thefootnote{}\footnote{#1}%
  \addtocounter{footnote}{-1}%
  \vspace{-1em}
  \endgroup
}
\author{Tom Hanika\inst{1,2}\putorcid{0000-0002-4918-6374}%
  \and Maximilian Marx\inst{3}\putorcid{0000-0003-1479-0341}%
  \and Gerd Stumme\inst{1,2}\putorcid{0000-0002-0570-7908}%
}

\institute{%
  Knowledge \& Data Engineering Group,
  University of Kassel, Germany
  \and
  ITeG,
  University of Kassel, Germany
  \and
  Center for Advancing Electronics Dresden (cfaed), TU~Dresden
  \\[0.5ex]
  \email{tom.hanika@cs.uni-kassel.de, maximilian.marx@tu-dresden.de,
    stumme@cs.uni-kassel.de}
}

\begin{document}
\maketitle
\blfootnote{Authors are given in alphabetical order. No priority in
  authorship is implied.}
\interfootnotelinepenalty=10000
\begin{abstract}
  Knowledge graphs have recently become the state-of-the-art tool for
  representing the diverse and complex knowledge of the
  world. Examples include the proprietary knowledge graphs of
  companies such as Google, Facebook, IBM, or Microsoft, but also
  freely available ones such as YAGO, DBpedia, and Wikidata. A
  distinguishing feature of Wikidata is that the knowledge is
  collaboratively edited and curated. While this greatly enhances the
  scope of Wikidata, it also makes it impossible for a single
  individual to grasp complex connections between properties or
  understand the global impact of edits in the graph. We apply Formal
  Concept Analysis to efficiently identify comprehensible implications
  that are implicitly present in the data. Although the complex
  structure of data modelling in Wikidata is not amenable to a direct
  approach, we overcome this limitation by extracting contextual
  representations of parts of Wikidata in a systematic fashion. We
  demonstrate the practical feasibility of our approach through
  several experiments and show that the results may lead to the
  discovery of interesting implicational knowledge. Besides providing
  a method for obtaining large real-world data sets for FCA, we sketch
  potential applications in offering semantic assistance for editing
  and curating Wikidata.
\end{abstract}

\keywords{Wikidata, Formal~Concept~Analysis, %Functional~Dependencies,
  Property~Dependencies, Implications}

% \begin{acks}
%   This work is partly supported by the \grantsponsor{gs:DFG}{German Research Foundation (DFG)}{}
%   within the \grantnum{gs:DFG}{Cluster of Excellence ``Center for Advancing Electronics Dresden'' (cfaed)}
%   and in \grantnum{gs:DFG}{Emmy Noether grant KR 4381/1-1 (DIAMOND)}.
%   % and in Emmy Noether grant KR 4381/1-1.
% \end{acks}

\section{Introduction}\label{sec:introduction}
The quest for the best digital structure to collect and curate
knowledge has been going on since the first appearances of knowledge
stores in the form of semantic networks and databases. The most
recent, and arguably so far most powerful, incarnation is the
\emph{knowledge graph}, as used by corporations like Facebook, Google,
Microsoft, IBM, and eBay. Among the freely available knowledge graphs,
Wikidata~\cite{wd-announcement,wikidata} stands out due to its free
and collaborative character: like Wikipedia, it is maintained by a
community of volunteers, adding \emph{items}, relating them using
\emph{properties} and \emph{values}, and backing up claims with
\emph{references}.
As of \DTMdate{2019-02-01}, Wikidata has 52,373,284 items and
676,854,559 statements using a total 5,592
properties. % But Wikidata also has a vibrant
% community: out of three million registered users, 20,793 have actively
% contributed to Wikidata throughout January~2019.
Altogether this
constitutes a gargantuan collection of factual data accessible to and
freely usable by everyone.

However, maintaining this large knowledge graph is not an easy task,
and retaining active editors poses an important challenge for the
community~\cite{wd-editor-evolution}: throughout the six years of
Wikidata's existence, Wikidata has amassed over three million
registered users, but only 20~thousand editors are still active. An
important step towards improving editor retention is to streamline the
editing process as much as possible. The largest fraction of edits is
made up of \emph{bot edits} (automated editing tools operated by
individual users) and \emph{batch edits} (i.e., mass edits done
through some tool specifically designed for certain types of
edits). These are primarily authored by seasoned editors not usually
susceptible to editor attrition~\cite{wd-editors-ontology}. In
contrast, casual editors typically do not use tools besides the
Wikidata web interface. Towards improving this editing experience, we
propose to extract implicational knowledge (“rules”) from Wikidata to
explicate to the editor the (potentially non-local) consequences of
editing a particular item's statements, similarly to the way
\emph{property
  constraints}\footnote{\url{https://www.wikidata.org/wiki/Help:Property_constraints_portal}}
are already used to highlight potentially conflicting or missing
data. Such rules must necessarily be easy to understand for editors
that are not already deeply familiar with Wikidata's data model and
ontological structure. Previous approaches have studied extracting
rules in the form of implications of first-order logic (FOL) is a
feasible approach to obtain interesting and relevant rules from
Wikidata.~\cite{Galarraga,rules-kg-learning} The expressive power of
FOL comes with a steep price, however: to understand such rules, one
needs to understand not only the syntax, but also advanced concepts
such as quantification over variables, and it seems far-fetched to
assume that the average Wikidata editor possesses such
understanding. We thus propose to use rules that are conceptually and
structurally simpler, and focus on extracting Horn implications of
\emph{propositional logic} (PL) from Wikidata, trading expressive
power for ease of understanding and simplicity of presentation.

While Formal Concept Analysis (FCA)~\cite{fca-book} provides techniques to
extract a sound and complete basis of PL implications (from which all
other implications can be inferred), applying these techniques to
Wikidata is not straightforward: A first hurdle is the sheer size of
Wikidata, necessitating the selection of subsets from which to extract
rules. Secondly, the intricate data model of Wikidata, while providing
much flexibility for expressing wildly different kinds of statements,
is not particularly amenable to a uniform approach to extracting
relevant information.

In this work, we tackle both issues by describing procedures
\begin{inparaenum}[i)]
\item for extracting, in a structured fashion, implicational knowledge
  for arbitrary subsets of properties, and
\item for deriving suitable sets of attributes from Wikidata
  statements, depending on the type of property.
\end{inparaenum}
We provide an implementation of these
procedures\footnote{\url{https://github.com/mmarx/wikidata-fca}}, and
while incorporating the extracted rules into the editing process is
out of scope for this paper, we nevertheless demonstrate that we are
able to obtain meaningful and interesting rules using our approach.

\section{Related Work}\label{sec:related-work}
In the realm of Wikidata, there have been two prominent applications
of FCA so far: The authors in~\cite{GH18:wikidata-dynamics} model and
predict the dynamic behaviour of knowledge graphs using lattice
structures, and~\cite{Lajus} attempts to determine obligatory
attributes for classes in Wikidata. A more general approach to
applying FCA to knowledge graphs was proposed in~\cite{fcakg}. A
current topic in knowledge bases (and in particular Wikidata) is Rule
mining. Several successful approaches to generating lists of FOL
rules, \eg{} in~\cite{rules-kg-learning,Galarraga} have been
proposed. This task is often connected to ranked lists of rules,
sometimes falsely denoted as recommendations, like
in~\cite{wd-recommender-comparison}, or completeness investigations
for knowledge graphs, like in~\cite{Galarraga17,TanonSRMW17}.

\section{Wikidata}\label{sec:wikidata-intro}
\paragraph{Data Model.}
Wikidata~\cite{wikidata} is the free and open Knowledge Graph of the
Wikimedia foundation. In Wikidata, \emph{statements} representing
knowledge are made using \emph{properties} that connect
\emph{entities} (either \emph{items} or other properties) to
\emph{values}, which, depending on the property, can be either items,
properties, \emph{data values} of one of a few data types, \eg{} URIs,
time points, globe coordinates, or textual data, or either of the two
special values \emph{unknown value} (\ie{} \emph{some} value exists,
but it is not known) and \emph{no value} (\ie{} it is known that there
is no such value).

\begin{example}\label{ex:wd-qualifiers-informal}
  Liz Taylor was married to Richard Burton. This fact is represented
  by a connection from item \lizTaylor{} to item \richardBurton{}
  using property \spouse.  But Taylor and Burton were married twice:
  once from 1964 to 1974, and then from 1983 to 1984.
\end{example}

To represent these facts, Wikidata enriches statements by adding
\emph{qualifiers}, pairs of properties and values, opting for
two “spouse” statements from Taylor to Burton with different
\startTime{} and \endTime{} qualifiers.

\paragraph{Metadata and Implicit Structure.}
Each statement carries metadata: \emph{references} track provenance of
statements, and the statement \emph{rank} can be used to deal with
conflicting or changing information. Besides \emph{normal} rank, there
are also \emph{preferred} and \emph{deprecated} statements. When
information changes, the most relevant statement is marked preferred,
\eg{} there are numerous statements for \population{} of \frankfurt{},
giving the population count at different times using the
\pointInTime{} qualifier, with the most recent estimate being
preferred. Deprecated statements are used for information that is no
longer valid (as opposed to simply being outdated), \eg{} when
the formal definition of a planet was changed by the International
Astronomical Union on \DTMdate{2006-09-13}, the statement that
\pluto{} is a \planet{} was marked deprecated, and an \endTime{}
qualifier with that date was added.

\begin{example}\label{ex:ranked-statements}
  We may write down these two statements in \emph{fact notation} as
  follows, where qualifiers and metadata such as the statement rank
  are written as an \emph{annotation} on the statement:

  \begin{align}
    \begin{split}\label{eq:sf-population}
      \population\statementBody{\frankfurt,
        736414}@\lbrack
      \determinationMethod\qualifierValue \\\estimation,
      \pointInTime\qualifierValue \text{\DTMdate{2016-12-31}},
      \texttt{rank}\qualifierValue \texttt{preferred}
      \rbrack
    \end{split}\\
    \begin{split}\label{eq:pluto-no-planet}
      \instanceOf\statementBody{\pluto,
        \planet}@\lbrack
      \endTime\qualifierValue \text{\DTMdate{2006-09-13}},\\
      \texttt{rank}\qualifierValue \texttt{deprecated}\rbrack
    \end{split}
  \end{align}
\end{example}

Further structure is given to the knowledge in Wikidata using
statements themselves: Wikidata contains a class hierarchy comprising
over 100,000 \emph{classes}, realised by the properties \instanceOf{}
(stating that an item is an \emph{instance} of a certain class) and
\subclassOf{}, which states some item $q$ is a \emph{subclass} of some
other class $q'$, \ie{} that all instances of $q$ are also instances
of $q'$.

\paragraph{Formalisation.}
Most models of graph-like structures do not fully capture the
peculiarities of Wikidata's data model. The generalised Property
Graphs~\cite{marpl}, however, have been proposed specifically to
capture Wikidata, and we thus phrase our formalisation in terms of a
\emph{multi-attributed relational structure}.

\begin{definition}\label{def:kg}
  Let $\mathcal{Q}$ be the set of Wikidata items, $\mathcal{P}$ be the
  set of Wikidata properties, and let $\mathcal{V}$ be the set of all
  possible data values. We denote by
  $\mathcal{E} \coloneqq \mathcal{Q} \cup \mathcal{P}$ the set of all
  entities, and define
  $\Delta \coloneqq \mathcal{E} \cup \mathcal{V}$. Now, the Wikidata
  knowledge graph is a map
  $\mathcal{W}\colon \mathcal{P} \to \powerset{\mathcal{E} \times \Delta
    \times \powerset{\mathcal{P} \times \Delta}}$ assigning to each
  property $p$ a ternary relation $\mathcal{W}(p)$, where a tuple
  $\tuple{s, v, a} \in \W{p}$ corresponds to a $p$-statement on $s$
  with value $v$ and annotation $a$.
\end{definition}

Thus, $\tuple{\Delta, \Family{\W{p}}{p \in \mathcal{P}}}$ is a
\emph{multi-attributed relational structure}, \ie{} a relational
structure in which every tuple is annotated with a set of pairs of
attributes and annotation values. While technically stored separately
on Wikidata, we will simply treat references and statement ranks as
annotations on the statements. In the following, we refer to the
Wikidata knowledge graph simply by $\mathcal{W}$. Furthermore, we
assume that deprecated statements and the special values \emph{unknown
  value} and \emph{no value} do not occur in $\mathcal{W}$. This is
done merely to avoid cluttering formulas by excluding these cases, and
comes without loss of generality.

\begin{example}\label{ex:wd-as-a-mars}
  Property \spouse{} is used to model marriages in
  Wikidata. Among others, $\W{\spouse}$ contains the two statements
  corresponding to the two marriages between Liz Taylor and Richard
  Burton from \cref{ex:wd-qualifiers-informal}:
  \begin{align}\label{eq:taylorSpouse1}
    \begin{split}
      \MoveEqLeft\tuple[\big]{\lizTaylor, \richardBurton, \\
        &\set{\tuple{\startTime, 1964},
          \tuple{\endTime, 1974}}}
    \end{split} \\
    \begin{split}\label{eq:taylorSpouse2}
      \MoveEqLeft\tuple[\big]{\lizTaylor, \richardBurton, \\
        &\set{\tuple{\startTime, 1983},
          \tuple{\endTime, 1984}}}
    \end{split}
  \end{align}
\end{example}

Next, we introduce some abbreviations for when we are not interested
in the whole structure of the knowledge graph.

\begin{definition}\label{def:projections}
  Let $R \subseteq S^{3}$ be a ternary relation over $S$. For
  $t = \tuple{s, o, a} \in S^{3}$, we denote by
  $\subject t \coloneqq s$ the \emph{subject} of $t$, by
  $\object t \coloneqq o$ the \emph{object} of $t$, and by
  $\annotation t \coloneqq a$ the \emph{annotation} of $t$,
  respectively. These extend to $R$ in the natural fashion:
  $\subject R \coloneqq \set{\subject t \given t \in R}$,
  $\object R \coloneqq \set{\object t \given t \in R}$, and
  $\annotation R \coloneqq \set{\annotation t \given t \in R}$,
  respectively. We indicate with \textasciicircum\ that a property is
  incident with an item as object:
  $\W{\inverseProperty{\spouse}}$ contains
  $\tuple{\richardBurton, \lizTaylor,
    \\\set{\tuple{\startTime, 1964}, \tuple{\endTime,
        1974}}}$.
\end{definition}

\section{Formal Contexts in Wikidata}
\label{sec:wikidata-context}
Building upon~\cref{def:kg}, we now recall basic notions from Formal
Concept Analysis and how they relate to the structure of
Wikidata. For a thorough introduction, we refer the reader
to~\cite{fca-book}. A \emph{formal context} is a triple
$\mathbb K = \tuple{G,M,I}$ where $G$ is a set of so-called
\emph{objects}, $M$ is a set of so-called \emph{attributes}, and
$I \subseteq G \times M$ is called the \emph{incidence relation}. This
relation gives rise to the definition of two \emph{derivation
  operations} traditionally sharing the same symbol:
$\cdot'\colon\powerset{G}\to\powerset{M}, A\mapsto A'\coloneqq
\set{m\in M\given\forall g\in A\colon \tuple{g,m}\in I}$ and
$\cdot'\colon\powerset{M}\to\powerset{G}, B\mapsto B'\coloneqq
\set{g\in G\given\forall m\in B\colon \tuple{g,m}\in I}$. Two sets
$A \subseteq G$ and $B \subseteq M'$ are called \emph{closed} in
$\mathbb K$ if $A = A''$ and $B = B''$, respectively.  A pair
$\tuple{A,B}$ satisfying $A'=B$ and $B'=A$, where $A\subseteq G$ and
$B\subseteq M$, is a \emph{formal concept}, the defining entity for
FCA. The set of all concepts is denoted by $\mathfrak{B}(\mathbb{K})$.
% , can be ordered using the order
% $\tuple{A,B}\leq \tuple{C,D}:\!\iff A\subseteq C$ for
% $\tuple{A,B},\tuple{C,D}\in\mathfrak{B}(\mathbb{K})$.
An \emph{attribute implication} is denoted by $X\to Y$, where $X,Y\subseteq
M$. We say $X\to Y$ is valid in $\mathbb{K}$ iff $X'\subseteq Y'$. % Furthermore,
% a \emph{model} of $X\to Y$ is a set $Z\subseteq M$, such that $X\subseteq
% Z\implies Y\subseteq Z$.
The set of all valid implications for $\mathbb{K}$ on
$M$ is called the \emph{attribute implicational theory}, denoted by
$\mathop{Th}_{M}(\mathbb{K})$. % The implicational theory
% $\mathop{Th}_{G}(\mathbb{K})$ on the set of objects is defined analogously.
In general, the theory of a formal context can be exponentially large
compared to the size of the context. Thus, one employs an implication
base, \ie{} a sound, complete, and non-redundant set of implications
from which the theory can be inferred. Among the various bases used in
FCA, the \emph{canonical base} $\mathcal{L}$ stands out due to its minimal
size~\cite{guigues1986famille}.

Reasoning using a canonical base is quite simple: for every attribute
set $X\subseteq M$, compute the closure $X^{\mathcal{L}}$, \ie{} apply
$\mathcal{L}$ to $X$ until the result is stable. This can be done in
time linear in the size of the canonical base~\cite{Beeri}. Thus,
entailment of an implication $X \to Y$ with respect to a base can be
decided in linear time: $X \to Y$ is entailed by $\mathcal{L}$ iff
$Y \subseteq X^{\mathcal{L}}$~\cite[Proposition 16]{GanterO16}. In
contrast, deciding entailment directly is \PTime-complete with respect
to the size of the implicational
theory~\cite{lp-complexity}. Computing the canonical base is thus a
more efficient way to decide entailment for multiple implications, as
the computational effort of computing a base gets amortised over the
entailment checks. An implication $X \to Y$ in a formal context
$\tuple{G, M, I}$ has \emph{support}\footnote{We use the definition
  for the support from FCA, which coincides with the definition of the
  support on \emph{valid} rules in association rules.}
$\supp(X \to Y) = \frac{\abs{X'}}{\abs{G}}$, \ie{} the relative number
of objects exhibiting the necessary attributes for the rule to be
applicable among all objects. A higher support implies that the
implication is more relevant to the whole domain of the
context. Nevertheless, a valid implication $X \to Y$ may have a
support of zero. % Such an implication witnesses that no
% \emph{counterexamples} exist in the domain, \ie{} that there is no
% object $g \in G$ with $X \subseteq \set{g}'$ but
% $Y \not \subseteq \set{g}'$.
% In the extreme case, a canonical base
% might contain only unsupported implications. Still, such a base allows
% checking entailment of a particular implication, \eg{} as a step
% towards algorithmic accountability.

\section{Property Theory}\label{sec:wikidata-exploration}
In the following, we employ these tools and techniques to obtain a
more accessible view on the Wikidata knowledge graph and how
properties therein depend on each other.
\citeauthor{ontologies-kgs}~\cite{ontologies-kgs} argues that
knowledge graphs are primarily characterised by three properties:
\begin{inparaenum}[i)]
\item normalised storage of information in small units,
\item representation of knowledge through the connections between
  these units, and
\item enrichment of the data with contextual knowledge.
\end{inparaenum}
In Wikidata, properties serve both as a mechanism to relate entities
to one another, as well as to provide contextual information on
statements through their use as qualifiers. Taking the structure and
usage of properties into account is thus crucial to any attempt of
extracting structured information from Wikidata. We introduce four
natural problem scenarios for selecting sets of properties from
Wikidata, each exploiting different aspects of the rich data model to
enhance the understanding of the data.

\subsection{Plain Incidence}\label{sec:plain-expl}
We start by constructing the formal context that has a chosen set
$\hat{\mathcal{P}} \subseteq \mathcal{P}$ as its attribute set and the
entity set $\mathcal{E}$ as the object set.

\begin{problem}\label{prob:plain}
  Given the Wikidata knowledge graph $\mathcal{W}$ and some subset
  $\hat{\mathcal{P}} \subseteq \mathcal{P}$, compute the canonical
  base for the implicational theory
  $\mathop{Th}_{\hat{P}}(\mathcal{E}, \hat{\mathcal{P}}, I^{\plain})$,
  where
  \begin{align}
    \tuple{e, \hat{p}} \in I^{\plain} :\!\iff e \in \subject
    \W{\hat{p}},\ \text{\ie}
  \end{align}
  an entity $e$ coincides with property $\hat{p}$ iff it occurs as a
  subject in some $\hat{p}$-statement.
\end{problem}

Although this is the most basic problem we present, with growing
$\hat{\mathcal{P}}$ it may quickly become computationally infeasible,
cf.~\cref{sec:evaluation}. More importantly, however, entities
occurring as objects are not taken into account: almost half of the
data in the knowledge graph is ignored, motivating the next definition.

\subsection{Directed Incidence}\label{sec:cont-expl}
We endue the set of properties $\mathcal{P}$ with two colours
$\set{\subject, \object}$ signifying whether an entity coincides
with the property as subject or as object in some
statement. % Hence, we choose some
% $\hat{\mathcal{P}} \subseteq \mathcal{P} \times \set{\subject,
%   \object}$.
\begin{problem}\label{prob:simp}
  Given $\mathcal{W}$ and some set
  $\hat{\mathcal{P}} \subseteq\mathcal{P}\times\{\subject,\object\}$
  of directed properties, compute the canonical base for
  $\mathop{Th}_{\hat{\mathcal{P}}}(\mathcal{E},\hat{\mathcal{P}},I^{\direction})$,
  where an entity $e$ coincides with $\hat{p}$ iff it occurs as
  subject or object (depending on the colour) of some $p$-statement:
  \begin{align}
    \begin{split}
    \tuple{e, \hat{p}} \in I^{\direction} :\!\iff &\big(\hat{p} =
    \tuple{p, \subject} \wedge e \in \subject \W{p}\big) \\
    \vee &\big(\hat{p} = \tuple{p, \object} \wedge e \in \object \W{p}\big).
    \end{split}
  \end{align}
\end{problem}

\begin{example}\label{ex:plain-insufficient}
  Let $\hat{\mathcal{P}}=\{\isMother{},\godparent{},\mother{}\}$ be
  the set of attributes and let
  $\mathcal{E}=\{\MileyCyrus{},\Victoria{},
  \NaomiWatts{},\\\AngelinaJolie{}\}$ be the set of objects. The
  corresponding formal context
  $\tuple{\mathcal{E},\hat{\mathcal{P}},I^{\direction}}$ (as extracted
  from Wikidata) is given by the following cross table:

  \begin{center}
    \begin{cxt}%
      \cxtName{Example}%
      \atr{\isMother{}}%
      \atr{\godparent{}}%
      \atr{\mother{}}%
      \obj{xxx}{\AngelinaJolie{}}
      \obj{.xx}{\MileyCyrus{}}
      \obj{x.x}{\NaomiWatts{}}
      \obj{xxx}{\Victoria{}}
    \end{cxt}
  \end{center}

  Observe that the only valid (non-trivial) implication (and hence
  sole constituent of the canonical base) is
  $\set{} \to \set{\mother}$: every entity has a mother.
\end{example}

\subsection{Qualified Incidence}\label{sec:qual-expl}
While \cref{prob:simp} captures \cref{ex:plain-insufficient}, it is
still insufficient to grasp the subtleties of \cref{ex:wd-as-a-mars},
since two statements differing only in their annotations are
indistinguishable. We thus include annotations into the colours of the
properties. For a property $p$, we define
$\Qualifiers{p} \coloneqq \bigcup_{t \in \W{p}}\annotation t$, the set
of all individual annotations occurring in statements for $p$.

\begin{problem}\label{prob:qual}
  For $\mathcal{W}$ and
  $\hat{\mathcal{P}} \subseteq \bigcup_{p \in \mathcal{P}} (\set{p}
  \times \set{\subject, \object} \times \Qualifiers{p})$, compute the
  canonical base for
  $\mathop{Th}_{\hat{\mathcal{P}}}(\mathcal{E},\hat{\mathcal{P}},I^{\qualifier})$,
  where
  \begin{align}
    \begin{split}
      \tuple{q, \hat{p}} \in I^{\qualifier} :\!\iff
      \big(\hat{p} = \tuple{p, \object, a} \wedge \exists t \in
      \W{p}.\ (\object t = q) \wedge (a \in \annotation t)\big) \\
      \vee \big(\hat{p} = \tuple{p, \subject, a} \wedge \exists t \in
      \W{p}.\ (\subject t = q) \wedge (a \in \annotation t)\big),\
      \text{\ie}
    \end{split}
   \end{align}
   an entity $e$ coincides with $\hat{p} = \tuple{p, d, a}$ iff it
   occurs as subject or object (depending on $d$) of some
   $p$-statement $t$, and the annotation $\annotation t$ of $t$
   includes $a$.
\end{problem}

\subsection{Classified Incidence}\label{sec:pred-expl}
Another natural approach to distinguishing properties is to consider
the classes that objects of the property are instances of: having a
\mother{} that is a \greekDeity{} is significantly different from one
that is merely a \human{}. We thus define for a property
$p \in \mathcal{P}$ the set of all classes that objects of
$p$-statements are instances of:
$\Instances{p} \coloneqq \{\object t \mid\, t\in\W{\instanceOf}, s \in
\W{p}, \object s = \subject t\}$.

\begin{problem}\label{prob:class}
  Given $\mathcal{W}$ and some set
  $\hat{\mathcal{P}} \subseteq \bigcup_{p \in P} (\set{p} \times
  \set{\subject, \object} \times \Instances{p})$, compute the
  canonical base for the implicational theory
  $\mathop{Th}_{\hat{\mathcal{P}}}(\mathcal{E},\hat{\mathcal{P}},I^{\instance})$,
  where
  \begin{align}
    \begin{split}
      \tuple{q, \hat{p}} &\in I^{\instance} :\,\iff \big(\hat{p} =
      \tuple{p, \subject, c} \wedge \exists s \in \W{p}.\\ \exists t
      &\in \W{\instanceOf}.\ (\subject s = q)\ \wedge
        (\object s = \subject t) \wedge (\object t = c)\big) \\
        \vee& \big(\hat{p} = \tuple{p, \object, c} \wedge \exists t \in
        \W{\instanceOf}.
        (q \in \object \W{p}) \\&\wedge (\subject t = q)
        \wedge (\object t = c)\big),\ \text{\ie{}}
    \end{split}
  \end{align}
  an entity $e$ coincides with $\hat{p} = \tuple{p, d, c}$ if there is
  a $p$-statement $t$ with subject (or object, respectively) $e$, and
  the object of $t$ is an instance of class $c$.
\end{problem}

\paragraph{Generalised Incidence.}\label{sec:gener-incid}
Any of the incidences so far may be combined into one generalised
incidence, since the point-wise union of two formal
contexts is again a formal context.
% : for two formal contexts
% $\mathbb{K}_{1} = (G_{1}, M_{1}, I_{1})$ and
% $\mathbb{K}_{2} = (G_{2}, M_{2}, I_{2})$, the union is
% $\mathbb{K}_{1} \cup \mathbb{K}_{2} \coloneqq (G_{1} \cup G_{2}, M_{1}
% \cup M_{2}, I_{1} \cup I_{2})$.
In the same spirit, one could investigate further incidences
emphasising other aspects of the Wikidata data model.

% Indeed, we consider such combinations
% in \cref{sec:wd-implications}, albeit with the mild restriction that
% no property may appear in more than one incidence. While the four
% incidences presented here are surely significant to the understanding
% of knowledge in Wikidata, this is by no means a complete list: one
% could imagine further incidences in the same spirit that emphasise
% other aspects of the Wikidata data model which might be more relevant
% to the subset of the knowledge graph in focus.

\section{Computations}
\label{sec:directsolv}
Solving~\cref{prob:plain,prob:simp,prob:qual,prob:class} should, in
theory, be straightforward: merely apply one of the two known
algorithms for computing the canonical base, or first compute some
other base like a \emph{direct base} from which one can deduce the
canonical base. However, for $\abs{\hat{\mathcal{P}}} \geq 200$,
computing the canonical base in reasonable time on affordable hardware
might already be impossible. Nonetheless, to demonstrate that it is
indeed possible to derive meaningful formal contexts from Wikidata
with the approaches of \cref{prob:plain,prob:simp}, we conducted
experiments with classical and more recent methods for computing
canonical bases. We discuss a range of different selections of
properties to illustrate these techniques.

\subsection{Data Sets}

\begin{table}[t]
  \centering
  \caption{Property selection in data sets}
  \begin{tabular}{@{}ll@{}}\toprule
    data set & properties in class (“Wikidata property for …”) \\\midrule
    \dataset{awards} & \propsAwards \\
    \dataset{family} & \propsFamily \\
    \dataset{math} & \propsMath \\
    \dataset{space} & \propsSpacecraft \\
    \dataset{time} & \propsTimeDuration \\\bottomrule
  \end{tabular}
  \label{tab:datasets-classes}
\end{table}

Our data sets were generated from the full Wikidata dump (in JSON
format) from
\DTMdate{2018-10-22}\footnote{\url{https://dumps.wikimedia.org/wikidatawiki/entities/20181022/}}. From
this dump, we have extracted different subsets by selecting fixed
sets of properties related to certain domains of interest (arbitrarily
chosen by the authors due to personal preference), which are
represented as classes of properties in Wikidata. Since Wikidata
comprises knowledge from a vast number of distinct and unconnected
domains, this is a natural simplification, as the properties used for,
\eg{} spacecraft are disjoint from those properties used for
mathematics. \Cref{tab:datasets-classes} describes which properties
were selected for which data set; \cref{tab:datasets-size} shows the
sizes of the various data sets. For each data set, we have extracted
formal contexts corresponding to the incidences of
\cref{prob:plain,prob:simp,prob:qual,prob:class}, respectively. In
generating these contexts, we ignore statements that are
\begin{inparaenum}[i)]
\item deprecated, since these are no longer considered valid,
\item have an unknown value, or
\item have no value.
\end{inparaenum}
All other statements contribute to populating the context according to
the corresponding incidence relation. Finally, we remove empty rows
and columns, since these do not influence implications.

For comparison, we have also included the data set \dataset{wiki44k}
as provided by~\cite{rules-kg-learning}, a small subset of simple
statements extracted from a Wikidata dump from
December~2014. Meanwhile, though, the usage of some properties on
Wikidata has changed, and, in particular, eight properties used in
this data set have since been deleted on Wikidata. Hence, we have also
generated \dataset{wiki44k-tr}, where these properties have
been replaced by their modern equivalents:
\begin{inparadesc}
\item[\brother, \sister:] replaced by \sibling,
\item[\grandparent:] replaced by \relative, with a \typeOfKinship{}
  qualifier with value \itemGrandparent{},
\item[\order:] replaced by \parentTaxon, where the object gets an
  additional \taxonRank{} statement with value \taxonOrder,
\item[\family:] replaced by \parentTaxon; where the object gets an
  additional \taxonRank{} statement with value \taxonFamily,
\item[\mainTypeGND, \administrativeEntity:]
  replaced by \instanceOf, and
\item[\languageFamily:] replaced by \subclassOf.
\end{inparadesc}
Both data sets were converted to JSON format and then processed analogously to
the other data sets. Data sets \dataset{wiki44k-2018} and
\dataset{wiki44k-2018-tr} are subsets of the 2018 dump obtained by dropping all
items and properties (and statements connecting those) not appearing in
\dataset{wiki44k} and \dataset{wiki44k-tr}, respectively.

\begin{table}
  \centering
  \caption{Size of data sets}
  \begin{tabular}{@{}llll@{}}\toprule
    data set & items & properties & statements \\\midrule
    \dataset{awards} & 429,207 & 27 & 892,723 \\
    \dataset{family} & 307,330 & 10 & 728,669 \\
    \dataset{math} & 36,913 & 45 & 84,255 \\
    \dataset{space} & 7,693 & 20 & 30,212 \\
    \dataset{time} & 216,865 & 9 & 219,803 \\
    \dataset{wiki44k} & 45,021 & 101 & 295,352 \\
    \dataset{wiki44k-2018} & 44,915 & 92 & 382,725 \\
    \dataset{wiki44k-tr} & 45,021 & 95 & 300,687 \\
    \dataset{wiki44k-2018-tr} & 44,919 & 94 & 384,700 \\\bottomrule
  \end{tabular}
  \label{tab:datasets-size}
\end{table}

\subsection{Empirical Evaluation}\label{sec:evaluation}
\Cref{tab:results} depicts the results of our computations
for~\cref{prob:plain}. The computation time for those results varied
from seconds in the case of \dataset{family} to approximately five
hours for the \dataset{wiki44-*} data sets. The program code for our
experiments was written in \emph{Clojure} and builds on the existing
\emph{conexp-clj}.\footnote{\url{https://github.com/exot/conexp-clj}}. A
\emph{GitHub}
repository\footnote{\url{https://github.com/wikiexploration/wikiexploration}}
holds data sets and computed results. We used a state-of-the-art
server system with two Intel\textregistered{} Xeon\textregistered{}
Gold 5122 CPUs and 768~GB RAM.

A first observation is that the canonical bases remain small in
relation to the data sets, even though the data is inherently noisy
and a considerable number of special items provokes peculiar
rules. This enhances the applicability of these bases. Secondly, we
observe that the number of supported rules varies strongly with
different data sets, \eg{} all rules are supported in the
\dataset{family} data set, whereas no rule in \dataset{time} is. Thus,
\dataset{time} does not admit a non-trivial propositional Horn logic
theory, but still, the canonical base comprises a plenitude of valid
rules. This could be due to incorrectly or insufficiently contributed
data, an unusually high number of exceptions in the domain, or even
due to a conscious design decision in the data modelling with respect
to the properties in \dataset{time}. Nonetheless, this canonical base
may still be used to validate that particular implications hold in the
domain.

\begin{table}[t]
  \centering
  \caption{Canonical bases for contexts in~\cref{prob:plain}}
  \begin{tabular}{@{}llll@{}}
    \toprule
    data set &density &$|\mathop{CanBase}(\cdot)|$ & \# supported\\\midrule
    \dataset{awards} &0.039&280&17 \\
    \dataset{family} &0.163&46&46  \\
    \dataset{math} &0.040&752&71  \\
    \dataset{space} &0.195&157&125 \\
    \dataset{time} &0.112&27&0  \\
    \dataset{wiki44k} &0.045&7,040&3,556 \\
    \dataset{wiki44k-2018} &0.053&8,179&5,550 \\
    \dataset{wiki44k-tr} &0.043&6,408&3,261\\
    \dataset{wiki44k-2018-tr} &0.053&9,422&6,641\\
    \bottomrule
  \end{tabular}
  \label{tab:results}
\end{table}

We now list some interesting, exemplary implications that we
discovered for \cref{prob:plain,prob:simp}. In the data set names,
\dataset{<data set>-0} corresponds to a context for \cref{prob:plain}, whereas
\dataset{<data set>-1} denotes a context for \cref{prob:simp}.
\begin{compactdesc}
\item[\dataset{awards-0}:]
  $\set{\nobelprizeid}\to\set{\awardReceived}$ is an implication
  supported by $0.2\%$ of the data set. While this is a reasonable
  implication, it is hardly surprising. Altogether, we obtained a set
  of 280 valid implications from which 17 are supported in the data
  set.
\item[\dataset{awards-1}:] We found the following implication, stating
  that everything that someone has been nominated for and that
  has an associated category for the award is also an award received
  by some entity, supported by $0.03\%$ of the data set:
  $\set{\isNominatedFor,\catRec}\\ \to \set{\isAwardReceived}$. Beyond
  this, implications from \dataset{awards-1} do not seem to shed more
  light on the set of investigated properties.
\item[\dataset{family-0}:] Here we computed a canonical base in which
  all implications are supported. For example, the implication
  $\set{\godparent,\partner}\to \set{\sibling}$ is supported by 7 out
  of 306,908 entities. It states that an entity that has a godparent
  and has a partner also has a sibling. However, this implication is
  not necessarily true for family relations.
\item[\dataset{family-1}:] With $0.03\%$ support, the implication
  $\set{\isFather,\isRelative,\\ \spouse}\to\set{\child}$ is
  unsurprising, but witnesses that the more general
  $\set{\isFather} \to \set{\child}$ has counterexamples in the data
  set: in fact, there are 1,634 non-fictional humans serving as
  counterexamples.
\item[\dataset{math-0}:] Among 752 implications, we discovered, with
  $0.01\%$ support, the implication
  $\set{\vFigure,\base}\to\set{\fPolytope}$. It is clearly helpful to
  obtain such rules when unfamiliar with polytope theory.
\item[\dataset{math-1}:] We observe a large number of rules relating
  mathematical identifiers, or at least using them. One such example,
  with support $0.02\%$, is the implication
  $\set{\vFigure,\mathworld, \isDualTo}\\\to
  \set{\fPolytope}$. Polytope theory is well represented in
  Wikidata. Other fields of mathematics are lacking data, however, and
  our hopes for numerous implications from diverse fields remained
  unfulfilled.
\item[\dataset{space-0}:] $4\%$ of the data set supports the
  implication $\set{\typeOfOrbit,\\\periapsis}\to
  \set{\apoapsis}$. From our point of view, there are several more
  elements of the computed canonical base contributing to a better
  understanding of the properties in this domain.
\item[\dataset{space-1}:] The rule (support $0.01\%$)
  $\set{\apoapsis, \isTypeOfOrbit}\\\to \set{\orbitalPeriod,
    \typeOfOrbit,\periapsis}$ states that all other relevant orbital
  facts are present for types of orbit with an apoapsis.
\item[\dataset{wiki44k-0}:] We find the rule
  $\set{\producer,\wdCountry} \to \set{\genre}$ with support $0.04\%$,
  stating that knowing the producer and the country of an item we can
  infer the genre associated to this item. Since this data collection
  was constructed with the explicit goal of having a very dense data
  set~\cite{Galarraga}, the probability for sparse counterexamples is low.
\end{compactdesc}

We claim that the implications discovered by us are readable and
comprehensible by humans, at least in cases where a user is familiar
with the domain of knowledge. Our experiments shed light on two
particular kinds of errors entailed in Wikidata. First, we found that
implications that should be valid, yet cannot be inferred from the
data set due to the presence of counterexamples. By incorporating a
background ontology, one may apply our method to identify missing
implications and therefore possible errors in Wikidata. These can be
fixed by editing statements or, in more serious cases, by introducing
new properties and deprecating (particular uses of) others. Secondly,
we observed valid implications that can easily be refuted by inquiring
a domain expert for a counterexample. Attribute exploration on parts
of Wikidata could be harnessed to systematically obtain such
counterexamples, although the sheer size of Wikidata requires a
collaborative exploration method.
At this point, we refrain from a thorough statistical evaluation, as,
\eg{} done in~\cite{rules-kg-learning}, since, from a logical
standpoint, the computed bases are not only sound and complete, but
also unique, methods such as cross-validation via, \eg{} embedding
models, are inappropriate to obtain meaningful measurements. Rather,
we focus on obtaining bases suitable for verifying implicational
assumptions on the data, as well as enhancing the comprehension of
properties by users of Wikidata.

\subsection{Limitations}\label{sec:limitations}
These experiments are subject to two limitations of our method:
\begin{inparaenum}[i)]
\item Already for a formalism as simple as horn PL rules, computing
  canonical bases is only feasible for small subdomains of
  Wikidata. This is hardly surprising, as recognising elements of the
  canonical base is \coNP-complete~\cite{Babin} (but becomes even
  harder for more expressive formalisms, if bases even exist). We
  investigate two known approaches for coping with this limitation in
  the next section.
\item The collaborative editing process is prone to introducing noise
  into the data, but the canonical base is sensitive to small changes
  and thus not well-suited for noisy data. This naturally leads to a
  weaker notion of base.
\end{inparaenum}
While technically also a limitation of the approach, we consciously
limit ourselves to implications in propositional Horn logic which
cannot express some of the rules obtained by more expressive
frameworks such as~\cite{rules-kg-learning,Galarraga}: computing
bases of implications for more expressive logics is computationally
infeasible at best, and impossible at worst. Moreover, propositional
Horn rules are arguably easier to grasp for untrained editors of
Wikidata than, \eg{} first-order rules.

\section{Association Rules and PAC Bases}\label{sec:bases-assoc-rules}
Various approaches to overcome the limitations stated in the last
section are known. A well-investigated and mature procedure is the
computation of \emph{association rules}~\cite{agrawal93}. While
closely related to implications in the FCA
sense~\cite{PasquierBTL99,stumme99conceptualknowledge,Zaki99}, they
were developed independently. For $X, Y \subseteq M$, the
\emph{confidence} of an association rule $X \to Y$ is
$\conf(X\to Y)\coloneqq\supp(X\cup Y)/\supp(X)$. The classical problem
then is to compute all association rules with minimum support
\texttt{minsup} and confidence at least \texttt{minconf} in a
domain. This usually leads to exponentially many rules. A plenitude of
extensions of association rules have been proposed, \eg{} different
kinds of support, head-confidence, etc. For our goal of obtaining a
base of implications, the most interesting extension was done
in~\cite{stummeAssociation}, relaying on the \emph{Luxenburger base of
  association rules}~\cite{luxenburger}. From this base one can infer
all valid association rules in the domain with respect to
\texttt{minsup} and \texttt{minconf}.

As a short empirical evaluation, we computed the Luxenburger base for
the data sets \dataset{awards}, \dataset{family}, \dataset{math},
\dataset{space}, and \dataset{time} for the setting of
\cref{prob:simp} with $\texttt{minsupp}=0.0001$ and
$\texttt{minconf}=0.6$. We list some interesting elements of these
bases:
\begin{inparadesc}
\item[awards (429,126 items, 38 properties):] The base has 28 rules,
  the rule $\set{\NMAW{},\nominatedFor{}} \to\set{\awardReceived{}}$
  has confidence $99.8\%$ and support $0.017\%$, with the sole
  counterexample \AlbertMaysles{}.\footnote{This error in the data has
    already been automatically flagged as a property constraint
    violation: \enquote{\emph{item requires statement constraint}: An
      entity with \emph{National Medal of Arts Winner ID} should also
      have a statement \emph{award received} \emph{National Medal of
        Arts}.}}
\item[family (306,908 items, 19 properties):] The rule
  $\set{\mother,\father}\to \set{\isChild}$ has a support $11.7\%$ and
  confidence $99.8\%$. The size of the base is 11,359.
\item[math (36,904 items, 63 properties):] Among the 482 rules is
  $\set{\inputSet{},\domain{}}\to \set{\codomain{}}$ which has support
  $0.024\%$ and confidence $88.8\%$, with the only counter example
  being \inverseF{}, due to a missing statement on this item.
\item[space (7,693 items, 30 properties):] The base has 666 elements. The rule
  $\set{\orbitalPeriod,\apoapsis}\to \set{\periapsis}$ is supported by
  $18\%$ of the data set and has a confidence of $99.9\%$ (note that
  this rule is not an element of the canonical basis).
\item[time (216,856 items, 15 properties):] The base has 4 elements. An
  interesting rule with support $0.03\%$ and confidence $97.4\%$ is
  $\set{\tempRangeEnd}\to \set{\tempRangeStart}$.
\end{inparadesc}
Except for \dataset{space}, the Luxenburger bases are of reasonable
size, \ie{} the size of the base is about 1\% of the number of
entities. We also computed the Luxenburger base for \dataset{wiki44k}
and \dataset{wiki44k-2018}, resulting in bases of 359,745 and 293,038
rules, respectively. While this is in sharp contrast to the positive
results above, there is a simple explanation for this effect: the
properties in our data sets were chosen from a common domain, leading
to a data set that includes items predominantly from this domain. The
\dataset{wiki44k} data set, however, was constructed by searching for
frequently-used (but not necessarily related)
properties~\cite{Galarraga}. This more arbitrary set of properties
results in a large number of rules crossing semantically independent
subsets of the property set.

A more recent approach is to employ the idea of PAC learning, as
introduced with the seminal paper by Valiant~\cite{Valiant84}. The
authors in~\cite{BorchmannHO17} present a procedure for retrieving
\emph{probably approximately correct implication bases} from a formal
context. For fixed probability $\delta$ and accuracy $\epsilon$, a PAC
basis can be computed in time polynomial in the size of the input
context and the size of the resulting basis. In contrast to
association rules, this approach still yields a canonical base (of
some approximation of the theory) consisting of (possibly)
unsupported, yet correct, implications.  A thorough investigation of
the applicability of PAC learning for Wikidata based formal contexts
requires its own research work and is out of scope for this
paper. Nevertheless, we provide first computational results in the
data repository.  Both of the methods discussed in this section are
more suitable for coping with the scale of Wikidata and the noise
inherently present in the data.

\section{Conclusion and Outlook}\label{sec:conclusion-outlook}
We have demonstrated in this work how to extract, in a structured
fashion, subsets of Wikidata and represent them as formal contexts,
thus opening a practically limitless source of real-world data to the
FCA community. This allows us (and indeed the whole community) to
apply the full range of tools and techniques from FCA to (arbitrary)
parts of Wikidata. Most importantly, we are now able to obtain
relevant and meaningful implications in a form that is readily
understood by untrained editors of Wikidata. Such rules are useful in
many different ways:
\begin{inparaenum}[i)]
\item they can further the understanding of knowledge implicit in
  Wikidata,
\item they can make explicit how editing a statement interacts with
  the implicational theory on properties,
\item they may highlight the need to edit further items to avoid
  introducing new counterexamples to valid rules, and
\item absence of expected rules serves as an indicator for errors
  present in the knowledge graph.
\end{inparaenum}
These qualities are highly desirable for streamlining the editing
experience on Wikidata, not only for casual editors, but also for
curators of Wikidata.

While we have shown that the direct computation of a canonical base is
feasible for small subsets of the data, this becomes infeasible as the
number of properties under consideration increases. We discussed two
different approaches to overcome this limitation: computing
Luxenburger bases of association rules, and computing a PAC basis,
both of which remain feasible on the scale of Wikidata.

A complementary approach would be to employ the well-known attribute
exploration algorithm to compute canonical bases for generalised
incidences over Wikidata. The key ingredient required for this is a
method to query Wikidata for possible counterexamples to proposed
implications, e.g., via the SPARQL endpoint. This enables Wikidata to
be used as an \emph{expert} for the exploration. Further expanding on
this, a collaborative exploration algorithm may employ both Wikidata
and human experts to stretch the boundaries of human knowledge.

Possible further directions for future work include
\begin{inparaenum}[i)]
\item a practical study on the usefulness of integrating implicational
  knowledge into the editing process,
\item integrating with completeness~\cite{rdf-completeness} tools such as
  COOL-WD~\cite{cool-wd} to ensure that only counterexamples above a certain
  completeness threshold are considered,
\item extending the structured approach to include further incidence
  relations adapted to other aspects of the Wikidata data model, such
  as grouping properties for quantities by intervals of their values,
  and
\item extend the approach to incorporate background knowledge given,
  \eg{} in the form of the MARPL rules~\cite{marpl} that have been
  proposed for ontological reasoning on Wikidata~\cite{sqid}.
\end{inparaenum}

\sloppy
\printbibliography

\end{document}